\documentclass[letterpaper, 10 pt, conference]{ieeeconf}  %

\IEEEoverridecommandlockouts                              %

\usepackage{amsmath} %
\usepackage{amssymb}  %
\usepackage{dsfont}
\usepackage{graphicx}

\usepackage{psfrag,graphicx,epsfig}
\usepackage{epstopdf}
\usepackage{xspace}
\usepackage{float}
\usepackage{placeins}
\usepackage{multirow}
\usepackage{pgf,tikz}
\usepackage{nowidow}
\usepackage{lineno}
\usepackage{xcolor}
\usepackage{arydshln} 

\usepackage{siunitx}
\usepackage{color}
\usepackage{flushend}
\usepackage{lineno}
\usepackage{algorithmic}%
\usepackage{url}
\usepackage{caption}
\usepackage{subcaption}
\usepackage{float}
\usepackage{hyperref}

\newcommand{\R}{\mathbb{R}}

\newcommand{\pyrobocop}{\textsc{PyRoboCOP}}
\newcommand{\adolc}{\textsf{ADOL-C}}
\newcommand{\ipopt}{\textsf{IPOPT}}
\newcommand{\casadi}{\textsf{CasADi}}
\newcommand{\pyomo}{\textsf{Pyomo}}

\newcommand{\mysub}[2]{[#1]_{#2}}
\newcommand{\setcompl}{{\cal L}}

\newcommand{\lb}[1]{\underline{#1}}
\newcommand{\ub}[1]{\overline{#1}}
\newcommand{\nfe}{N_e}

\newcommand{\setnfe}{{\cal N}_e}

\title{\LARGE \bf
\pyrobocop : \underline{Py}thon-based \underline{Robo}tic \underline{C}ontrol \& \underline{O}ptimization \underline{P}ackage 
for Manipulation 
}

\author{Arvind U. Raghunathan$^{1}$, Devesh K. Jha$^{1}$and Diego Romeres$^{1}$%
\thanks{$^1$All authors are with Mitsubishi Electric Research Laboratories (MERL), Cambridge, MA 02139. Email--{ \tt\small \{raghunathan,jha, romeres\}@merl.com}}%
}

\begin{document}

\maketitle
\thispagestyle{empty}
\pagestyle{empty}

\begin{abstract}
\pyrobocop\, is a Python-based package for control, optimization and estimation of robotic systems described by nonlinear Differential Algebraic Equations (DAEs).  In particular, the package can handle systems with contacts that are described by complementarity constraints and provides a general framework for specifying obstacle avoidance constraints. The package performs direct transcription of the DAEs into a set of nonlinear equations by performing orthogonal collocation on finite elements.    \pyrobocop\, provides automatic reformulation of the complementarity constraints that are tractable to NLP solvers to perform optimization of robotic systems. The package is interfaced with \adolc\,~\cite{ADOLC} for obtaining sparse derivatives by automatic differentiation and \ipopt\,~\cite{IPOPT} for performing optimization. We evaluate \pyrobocop\ on several manipulation problems for control and estimation. %
\end{abstract}

\section{Introduction}\label{sec:introduction}
Most manipulation applications are characterized by presence of constrained environments while dealing with challenging underlying phenomena like unilateral contacts, frictional contacts, impact and deformation~\cite{mason2018toward}. With this understanding, we present a python-based robotic control and optimization package (called \pyrobocop) that allows solution to a large class of mathematical programs with nonlinear and complementarity constraints. The current paper and package only considers systems which can be represented by DAEs. Integration with physics engines is left as a future work as that requires additional development.%

Contact-rich robotic manipulation tasks could be modeled as complementarity systems. Obtaining a feasible, let alone an optimal trajectory, can be challenging for such systems.  An effective integration of the high-level trajectory planning in configuration space with physics-based dynamics is necessary in order to obtain optimal performance of such robotic systems. To the best of author's knowledge, none of the existing python-based open-source optimization packages can provide support for trajectory optimization with support for complementarity constraints that arise from contact-rich manipulation and the easy specification of obstacle avoidance constraints. Such optimization capability is, however, highly desirable to allow easy solution to optimization problems for a large-class of contact-rich robotic systems. 

In this paper, we present \pyrobocop\,, a lightweight but powerful Python-based package for control and optimization of robotic systems. The formulation in \pyrobocop\ allows us to handle contact and collision avoidance in an unified manner. \pyrobocop\ uses \adolc\,~\cite{ADOLC} and \ipopt\,~\cite{IPOPT} at its backend for automatic differentiation and optimization respectively.
The main features of the package~are:
\begin{itemize}
    \item Contact modeling by complementarity constraints
    \item Obstacle avoidance modeling by complementarity constraints
    \item Automatic differentiation for sparse derivatives
    \item Support for minimum time problems
    \item Support for optimization over fixed mode sequence problems with unknown sequence time horizons
    \item Support for parameter estimation in linear complementarity systems.
\end{itemize}
The features described above should convince the reader that \pyrobocop\, addresses lots of important optimization problems for manipulation systems. By bringing together \adolc\,~\cite{ADOLC} and \ipopt\,~\cite{IPOPT} we believe that \pyrobocop\ would be very useful for real-time model-based control of manipulation systems. Codes and instructions for installing and using \pyrobocop\ could be found in \cite{pyrobocop_software} under a research only license. \\
\textbf{Contributions.} The main contributions of the paper are:
\begin{enumerate}
    \item We present a python-based package for optimization and control of robotic systems with contact and collision constraints.
    \item We evaluate our proposed package, \pyrobocop\ , over a range of different manipulation systems for control and estimation.
\end{enumerate}
For the sake of space, we do not report the formulation of collision avoidance with complementarity constraints. The interest reader can refer to the arXiv version of the work~\cite{raghunathan2021pyrobocop}.

\section{Related Work}~\label{sec:related_work}
Our work is closely related to various optimization techniques proposed to solve contact-implicit trajectory optimization. Some related examples could be found in~\cite{manchester2020variational, patel2019contact, erez2012trajectory, mordatch2012contact, mordatch2012discovery, YuntGlocker05,YuntGlocker07,Yunt11}. In a more general setting, our work is related to trajectory optimization in the presence of non-differentiable constraints. These problems are common in systems with constraints like non-penetrability~\cite{posa2014direct}, minimum distance (e.g., in collision avoidance)~\cite{zhang2018autonomous}, or in some cases robustness constraints~\cite{kolaric2020local}. 

Some of the existing open-source software for dynamic optimization are Optimica~\cite{Optimica2010},  ACADO Toolkit~\cite{ACADO2011}, TACO~\cite{TACO2013}, pyomo.dae~\cite{PYOMO.DAE2018}, Drake~\cite{drake} and CasADi~\cite{CASADI2019}.   All of the cited software leverage automatic differentiation to provide the interfaced NLP solvers with first and second-order derivatives.  However, these software do not provide convenient interfaces for handling complementarity constraints and obstacle-avoidance which are key requirements in robotic applications. More recently, some packages have been proposed to perform contact-rich tasks in robotics~\cite{9196673}. However, the solver proposed in~\cite{9196673} uses DDP-based~\cite{murray1979constrained} techniques which suffer from sub-optimality and difficulty in constraint satisfaction. Compared to most of the other techniques in open literature, \pyrobocop\ implements different formulations for handling complementarity constraints using NLP solvers~robustly.

The optimization method presented in our work is most closely related to the direct trajectory optimization method for contact-rich systems earlier presented in~\cite{YuntGlocker05,YuntGlocker07,Yunt11}. The authors pose contact dynamics as a measure differential inclusion and employ an augmented Lagrangian to solve the resulting complementarity constrained optimization problem. \cite{posa2014direct} handle the complementarity constraint by relaxing to an inequality and solving using an active-set solver. In an analogous manner, other optimization packages like \casadi\ and \pyomo\ can also be extended for the solution to trajectory optimization in presence of complementarity constraints through a similar reformulation of the complementarity constraints. We provide an adaptive approach for relaxing the complementarity constraints. Further, we also provide a novel formulation for trajectory optimization in the presence of minimum distance constraints for collision avoidance. To the best of our knowledge, there is no other existing open-source, python-based optimization package that can handle constraints arising due to frictional contact interaction and collision avoidance. %

\section{Problem Description} \label{sec:probdesc}

\pyrobocop\, solves the dynamic optimization problem
\begin{subequations}
\begin{align}
    \min\limits_{x,y,u,p}        &\, \int\limits_{t_0}^{t_f} c(x(t),y(t),u(t),p) dt + \phi(x(t_f),p)\label{dynopt:obj} \\
    \text{s.t.} &\, f(\dot{x}(t),x(t),y(t),u(t),p) = 0,\, x(t_0) = \widehat{x} \label{dynopt:daeqn} \\
                &\, (\mysub{y(t)}{\sigma_{l,1}} - \nu_{l,1}) (\mysub{y(t)}{\sigma_{l,2}} - \nu_{l,2}) = 0 \,\forall\, l \in \setcompl \label{dynopt:ceqn} \\
                &\, \lb{x} \leq x(t) \leq \ub{x}, \lb{y} \leq y(t) \leq \ub{y}, \lb{u} \leq u(t) \leq \ub{u} \label{dynopt:bnds}
\end{align}\label{dynopt}
\end{subequations}
where $x(t) \in \R^{n_x}$, $y(t) \in \R^{n_y}$, $u(t) \in \R^{n_u}$, $\dot{x}(t) \in \R^{n_x}$, $p \in \R^{n_p}$  are the differential, algebraic, control, time derivative of differential variables and time-invariant parameters respectively. The function $\phi : \R^{n_x+n_p} \rightarrow \R$ represents Mayer-type objective function~\cite{BrysonHo} term and is not a function of the entire trajectory.  In addition, $\underline{x}, \overline{x}$, $\underline{y}, \overline{y}$, $\underline{u}, \overline{u}$ are the lower and upper bounds on the differential, algebraic and control variables. The initial condition for the differential variables is $\widehat{x}$.    
Constraints~\eqref{dynopt:daeqn}-\eqref{dynopt:ceqn} are the Differential Algebraic Equations (DAEs) modeling the dynamics of the system with $f : \R^{2n_x+n_y+n_u} \rightarrow \R^{n_x+n_y-n_c}$ with $n_c = |\setcompl|$. 
Each $l \in \setcompl$ defines a pair of indices $\sigma_{l,1},\sigma_{l,2} \in \{1,\ldots,n_y\}$ that specifies the complementarity constraint between the algebraic variables $\mysub{y(t)}{\sigma_{l,1}}$ and $\mysub{y(t)}{\sigma_{l,2}}$.  In~\eqref{dynopt:ceqn} $\nu_{l,1}, \nu_{l,2}$ correspond to either the lower or upper bounds on the corresponding algebraic variables.  For example, if they are set respectively to the lower and upper bounds of corresponding algebraic variables then~\eqref{dynopt:ceqn} in combination with the bounds~\eqref{dynopt:bnds} model the complementarity constraint
\[
0 \leq \mysub{ y(t) - \lb{y} }{\sigma_{l,1}} \perp 
\mysub{ \ub{y} - y(t) }{\sigma_{l,2}} \geq 0.
\]

The dynamic optimization problem in~\eqref{dynopt} is transcripted to a NonLinear Program (NLP) using the Implicit Euler time-stepping scheme.  The time interval $[t_0,t_f]$ is discretized into $\nfe$ finite elements of width $h_i$ such that $\sum_{i = 1}^{\nfe}h_i = t_f - t_0$. Let $t_i = t_0 + \sum_{i' \leq i} h_{i'}$ denote the ending time of the finite elements $i$.  
The NLP that results is
\begin{subequations}
\begin{align}
    \min        &\, \sum\limits_{i=1}^{\nfe} h_i c(x_{i},y_{i},u_{i-1}) + \phi(x_{\nfe},p) \label{nlp:obj} \\
    \text{s.t.} &\, f(\dot{x}_{i},x_{i},y_{i},u_{i-1}) = 0,\, x_{0} = \widehat{x} \label{nlp:daeqn} \\
                &\, (\mysub{y_{i}}{\sigma_{l,1}} - \nu_{l,1}) (\mysub{y_{i}}{\sigma_{l,2}} - \nu_{l,2}) = 0 \,\forall\, l \in \setcompl \label{nlp:ceqn} \\
                &\, \lb{x} \leq x_{i} \leq \ub{x}, \lb{y} \leq y_{i} \leq \ub{y}, \lb{u} \leq u_{i} \leq \ub{u} \label{nlp:bnds} \\
                &\, x_{i+1} = x_i + h_i \dot{x}_i \label{nlp:xcont} 
\end{align}\label{nlp}
\end{subequations}
where the decision variables in~\eqref{nlp} are $x_{i}$, $\dot{x}_{i}$, $y_{i}$ and $u_{i}$. The subscript $i$ approximates the value of the corresponding variable at time $t_i$. The constraints~\eqref{nlp:daeqn}-\eqref{nlp:ceqn} are imposed for $i \in \setnfe$. The constraint in~\eqref{nlp:xcont} models Implicit Euler time-stepping scheme and is imposed for $i \in \setnfe \setminus \{\nfe\}$.  If complementarity constraints are present then~\eqref{nlp} is an instance of a Mathematical Program with Complementarity Constraints (MPCC).

MPCCs are well known to fail the standard Constraint Qualification (CQ) such as the Mangasarian Fromovitz CQ (MFCQ), see \cite{luo1996mathematical}.  Hence, solution of MPCCs has warranted careful handling of the complementarity constraints when used in Interior Point Methods for NLP (IPM-NLP) using relaxation~\cite{ipoptc,twosidedipm} or penalty formulations~\cite{knitroc}.  In the case of active set methods, the robust solution of MPCC relies on special mechanism such as the elastic mode \cite{elasticmode}.  

\pyrobocop\, implements two possible relaxation schemes for complementarity constraints 
\begin{subequations}
\begin{align}
    \alpha_l (\mysub{y_{i}}{\sigma_{l,1}} - \nu_{l,1}) (\mysub{y_{i}}{\sigma_{l,2}} - \nu_{l,2}) \leq \delta \,\forall\, l \in \setcompl \label{ceqn:opt1} \\
  \sum\limits_{l \in \setcompl} \alpha_l (\mysub{y_{i}}{\sigma_{l,1}} - \nu_{l,1}) (\mysub{y_{i}}{\sigma_{l,2}} - \nu_{l,2}) \leq \delta \label{ceqn:opt2}
\end{align}
\end{subequations}
where $\alpha_l = 1$ if the involved bounds ($\nu_{l,1},\nu_{l,2}$) are either both lower or both upper bounds. If one of the bounds is a lower bound and other is an upper bound then $\alpha_l$ is set to $-1$. Note that the choice of $\alpha_l$ ensures that the resulting product is nonnegative whenever~\eqref{nlp:bnds} are satisfied.  The first approach relaxes each complementarity constraint by a positive parameter $\delta$~\cite{ipoptc} while the second approach imposes the relaxation on the summation of all the complementarity constraints over a finite element $i$~\cite{elasticmode}.
\begin{figure}[!ht]
    \centering
    \includegraphics[width=.47\textwidth]{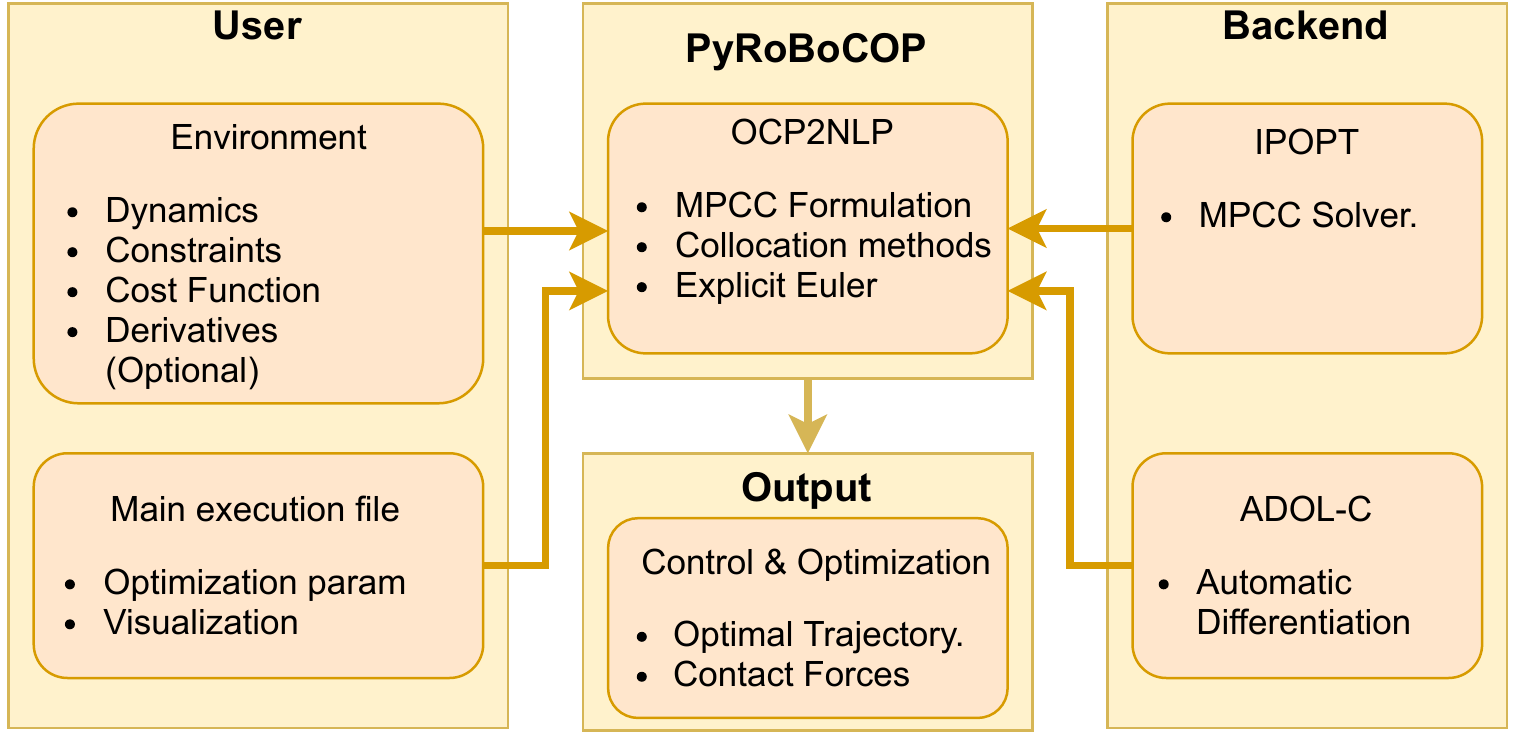}
    \caption{Workflow in \pyrobocop. The dynamics provided by the user to create a MPCC which is then optimized using \ipopt\ and the gradients are evaluated using automatic differentiation via \adolc. }
    \label{fig:pyrobocop}
\end{figure}
In addition, we also have flexibility to keep the $\delta$ fixed to a constant parameter through out the optimization or link this with the barrier parameter in IPM-NLP~\cite{ipoptc,twosidedipm}.

\section{Software Description}\label{sec:software_description}

Figure~\ref{fig:pyrobocop} provides a high-level summary of the flow of control in \pyrobocop. Detailed descriptions on the software API and classes are available in the Software Description in~\cite{pyrobocop_software}. A user provided class specifies the dynamic optimization problem~\eqref{dynopt}. This is also briefly described in Figure~\ref{fig:pyrobocop}. The user needs to provide the equality constraints for the dynamical system. These constraints could include the dynamics information for the system, the bounds on the system state and inputs, and information about complementarity constraints, if any. Furthermore, a user needs to provide the objective function, and also has the option to provide derivative information (note the derivative information is optional).  \pyrobocop\, expects the user provided class to implement the following methods in order to formulate an MPCC (or NLP) (also shown in Figure~\ref{fig:pyrobocop}).
\begin{itemize}
    \item \texttt{get\_info}: Returns information on~\eqref{dynopt} including $n_d$, $n_a$, $n_u$, $|\setcompl|$, $n_p$, $N_e$, $h_i$.
    \item \texttt{bounds}: Returns the lower and upper bounds on the variables $x(t)$, $\dot{x}(t)$, $y(t)$, $u(t)$ at a time instant $t$.
    \item \texttt{initialcondition}: Returns the initial conditions for the variables $x(t_0)$, i.e. values of the differential variables at initial time instant $t_0$.
    \item \texttt{initialpoint}: Returns the initial guess for the variables $x(t)$, $\dot{x}(t)$, $y(t)$, $u(t)$ at a time instant $t$. This initial guess is passed to the NLP solver.
    \item \texttt{objective}: Implements method to evaluate and return  $c(x(t),y(t),u(t),p)$ at a time instant $t$.
    \item \texttt{constraint}: Implements method to evaluate and return $(f(x(t),\dot{x}(t),y(t),u(t),p)$ at a time instant $t$.
\end{itemize}
We provide a description of optional methods that are expected if certain specified conditions are satisfied.
\begin{itemize}
    \item \texttt{bounds\_finaltime}: Returns the bounds on the variables $x(t_f)$ at the final time. This method allows to specify a final time condition on a subset or all of the differential variables.
    \item \texttt{bounds\_params}: Returns information on lower and upper bounds on the parameters $p$. This method must be implemented if $n_p > 0$.
    \item \texttt{initialpoint\_params}: Returns the initial guess for the parameters $p$. This method  must be implemented if $n_p > 0$. This initial guess is passed to the NLP solver.
    \item \texttt{get\_complementarity\_info}: Returns information on the complementarity constraints in~\eqref{dynopt} i.e. $\setcompl$ and also information on whether the lower or upper bound is involved in the complementarity constraint. This method must be implemented if $\setcompl \neq \emptyset$.
    \item \texttt{objective\_mayer}: Implements method to evaluate and return $\phi(x(t_f),p)$.
    \item \texttt{get\_objects\_info}: Returns the information on number of objects $n_O$, flags to indicate if these obstacles are static or dynamic and the number of vertices $n_{vi}$ for the polytope bounding the objects.
    \item \texttt{get\_object\_vertices}: Implements and returns the matrix $V_i(x(t),y(t)) \in \R^{3 \times n_{vi}}$ representing the vertices of the polytope bounding the objects. This method is called only when \texttt{get\_objects\_info} is implemented and $n_O > 0$.
\end{itemize}

\pyrobocop\, is interfaced with \adolc\,~\cite{ADOLC} to compute derivatives (see the Backend block in Figure~\ref{fig:pyrobocop}).  Note that the \adolc\, can also provide the sparsity pattern of the constraint Jacobian and Hessian of the Lagrangian. As mentioned earlier, the exploitation of sparsity in computations of the NLP is critical to solve large problems. To provide derivatives \pyrobocop\, used \adolc\, to set up tapes~\cite{ADOLC} for evaluating: (i) the objective~\eqref{nlp:obj}, (ii) constraints including the DAE~\eqref{nlp:daeqn} and a reformulation of~\eqref{nlp:ceqn}, and (iii) the Hessian of the Lagrangian of the NLP~\eqref{nlp}.  The set-up of the tape is done prior to passing control to the NLP solver.  The advantage of this approach is that evaluation of~\eqref{nlp:obj},~\eqref{nlp:daeqn} are now C-function calls instead of Python-function calls.  This considerably reduced the time spent in function evaluations for the NLP solver. As shown in Figure~\ref{fig:pyrobocop}, \pyrobocop\ uses \ipopt\ as the optimization solver. The choice of the solver agrees with our desire of having an open source software package, for this reason \ipopt\ is chosen over other competitors like SNOPT~\cite{gill2005snopt} and filterSQP~\cite{fletcher2002nonlinear}.

The user has flexibility in specifying how the complementarity constraints are solved.  The choices are: (i)~\eqref{ceqn:opt1} with $\delta$ fixed, (ii)~\eqref{ceqn:opt2} with $\delta$ fixed, (iii)~\eqref{ceqn:opt1} with $\delta$ set equal to the interior point barrier parameter, (iv)~\eqref{ceqn:opt2} with $\delta$ set equal to the interior point barrier parameter, and (v) the objective function is appended with complementarity terms as $\sum_{i=1}^{N_e}\sum\limits_{l \in \setcompl} \alpha_l (\mysub{y_{i}}{\sigma_{l,1}} - \nu_{l,1}) (\mysub{y_{i}}{\sigma_{l,2}} - \nu_{l,2})$.  The convergence behavior of formulations can be quite different and we provide these implementations so the user can choose one that works best for the problem at hand. As empirical guidelines, we recommend options (iii) and (iv) since they enforce the complementarity constraints in a gradual manner as the algorithm converges to a solution.
\section{Numerical results}\label{sec:results}
In this section, we test \pyrobocop\ in several robotic simulations providing solutions to trajectory optimization problems including several systems with complementarity constraints. To foster reproducibility, the source code for each of the following examples is available in \cite{pyrobocop_software}.%

\begin{figure*}[!h]
     \centering
     \begin{subfigure}[b]{0.25\textwidth}
         \centering
         \includegraphics[width=\textwidth]{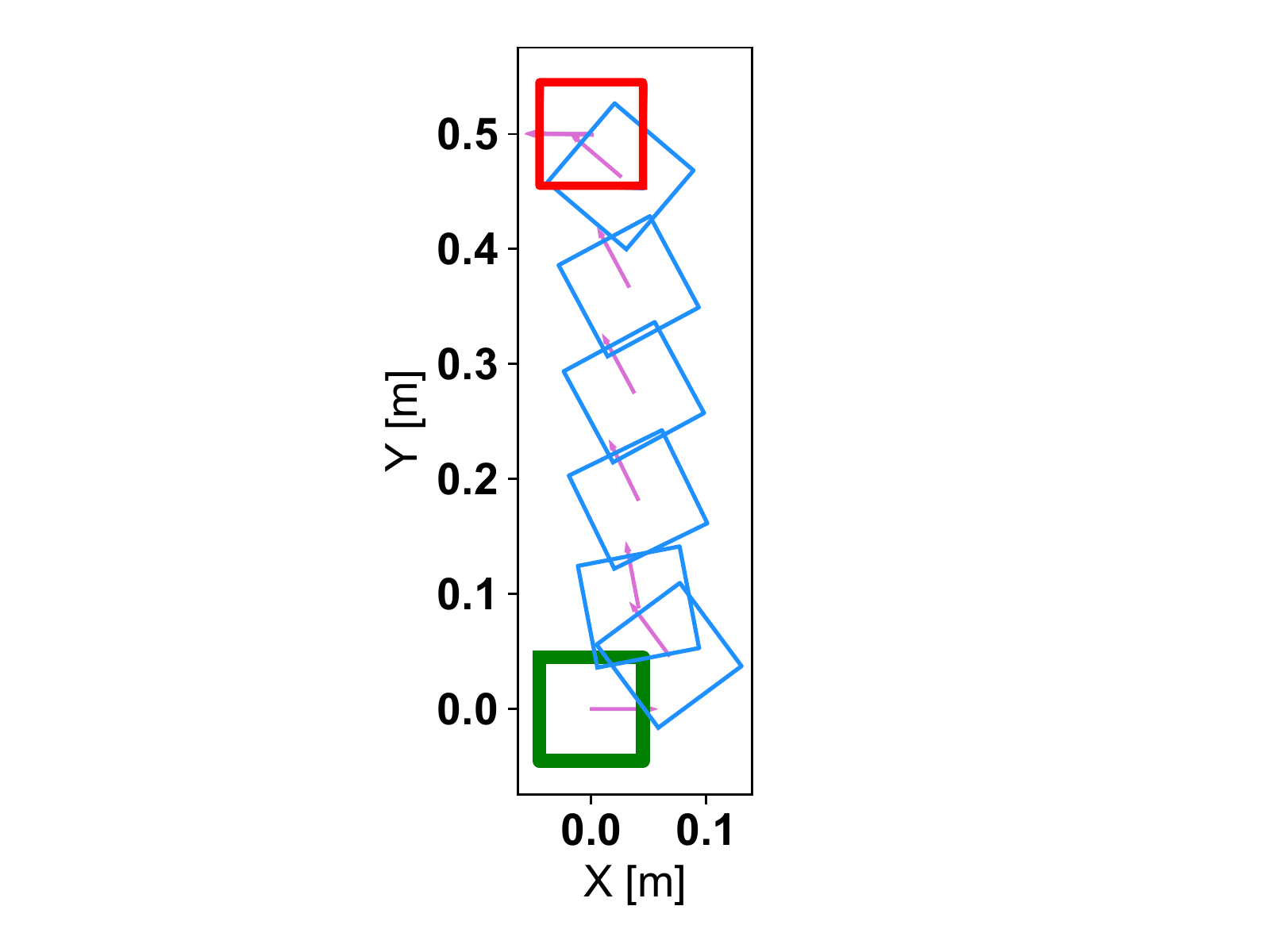}
		\vskip -0.05in
		\subcaption{Trajectory for 
		$\mathbf{x_g}=(0,0.5,\pi)$}\label{fig:pushing_ex1}
     \end{subfigure}
     \hfill
     \begin{subfigure}[b]{0.24\textwidth}
         \centering
         \includegraphics[width=\textwidth]{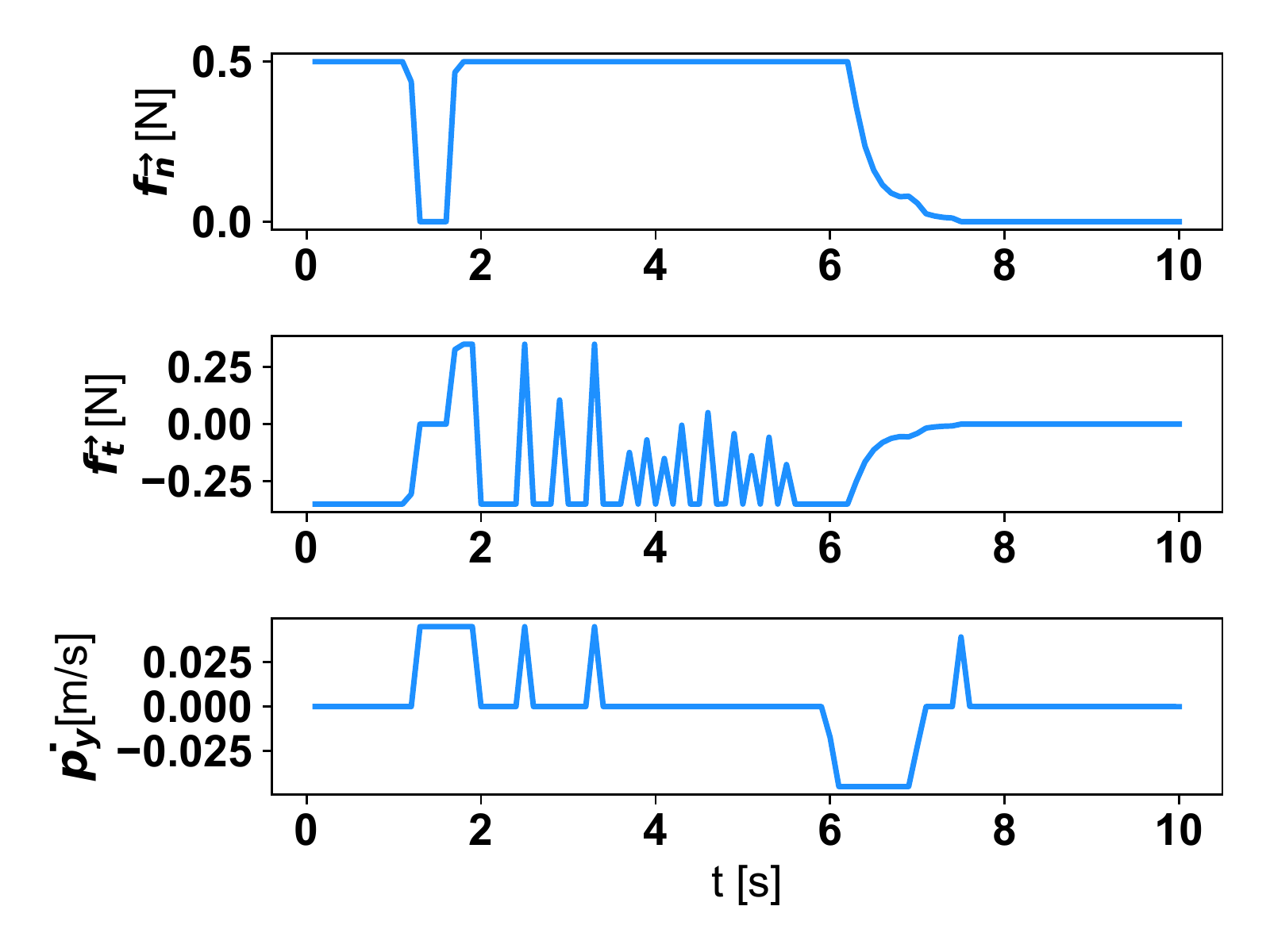}
		\vskip -0.05in
		\subcaption{Optimal Controls}\label{fig:pushing_ex1_input}
     \end{subfigure}
     \hfill
          \centering
     \begin{subfigure}[b]{0.24\textwidth}
         \centering
         \includegraphics[width=\textwidth]{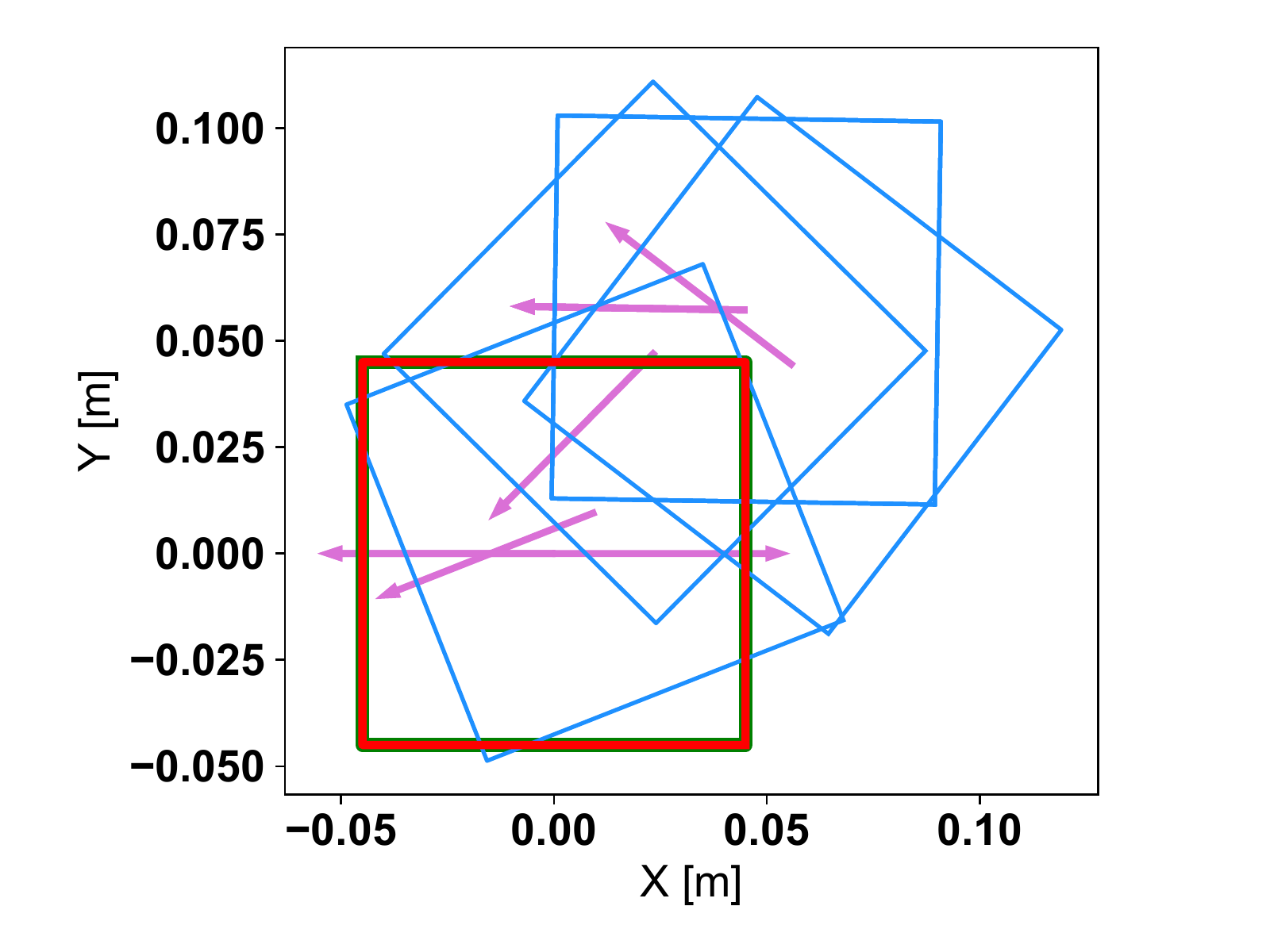}
		\vskip -0.05in
		\subcaption{Trajectory for $\mathbf{x_g}=(0,0,\pi)$}\label{fig:pushing_ex2}
     \end{subfigure}
     \hfill
     \begin{subfigure}[b]{0.24\textwidth}
         \centering
         \includegraphics[width=\textwidth]{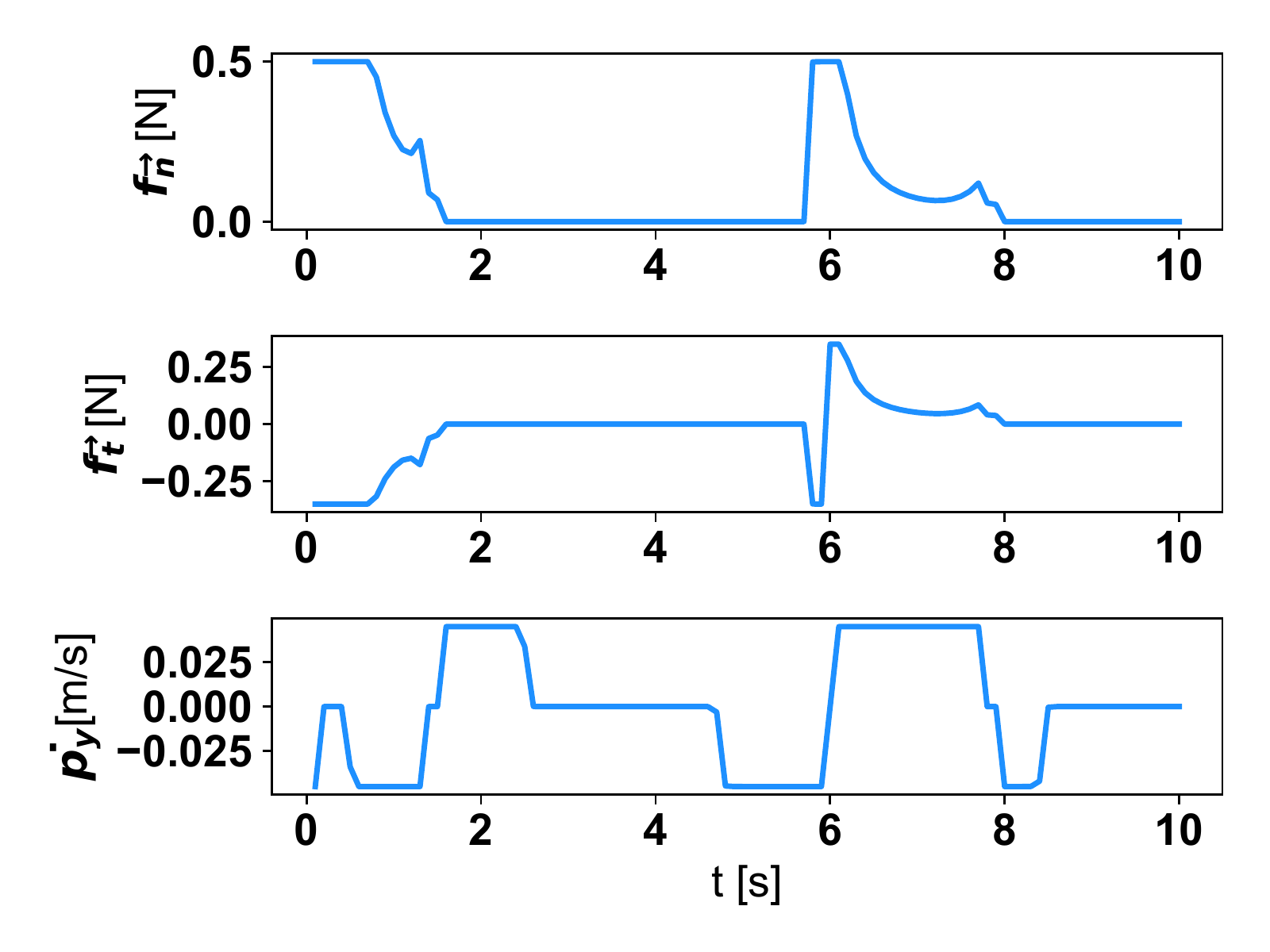}
		\vskip -0.05in
		\subcaption{Optimal Controls}\label{fig:pushing_ex2_input}
     \end{subfigure}
     \caption{Optimal pushing sequences and control inputs obtained by solving the MPCC for two different goal conditions. The switching sequence between sticking and slipping contact formation could be visualized by the trajectory of $\dot{p}_y$. The pusher maintains a sticking contact with the slider when $\dot{p}_y=0$. For clarity, we show only few frames of the trajectory in Fig.~\ref{fig:pushing_ex2}.}
	\label{fig:example_pushing}
\end{figure*}
\begin{figure}
    \centering
    \includegraphics[scale=0.25]{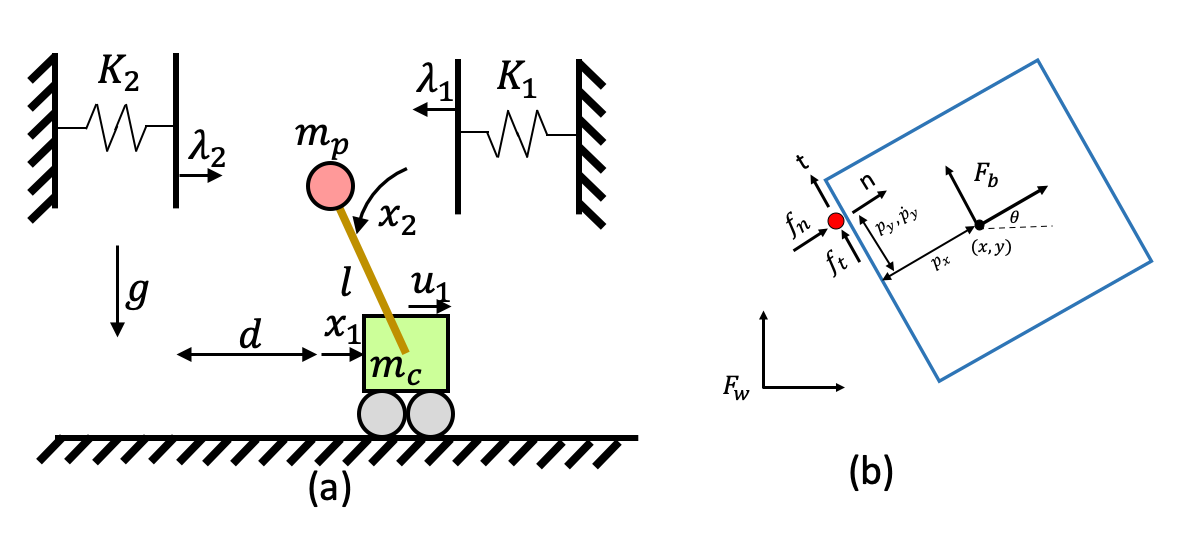}
    \caption{A schematic of cartpole with softwalls and planar pusher-slider system we study in this paper.}
    \label{fig:pushing_analytical}
\end{figure}

\subsection{Planar Pushing}\label{subsec:planar_pushing}

In this section, we show some results for planar pushing. We briefly describe the dynamic model here .For more detailed description of the pushing model, readers are referred to~\cite{hogan2018reactive} and \cite{pmlr-v87-bauza18a}. A schematic for a pusher-slider system is shown in Figure~\ref{fig:pushing_analytical}. The pusher interacts with the slider by exerting forces in the normal and tangential direction denoted by $f_{\overrightarrow{n}}$, $f_{\overrightarrow{t}}$ (as shown in Figure~\ref{fig:pushing_analytical}) as well as a torque $\tau$ about the center of the mass of the object. Assuming quasi-static interaction, the limit surface~\cite{GOYAL1991307} defines an invertible relationship between applied wrench $\mathbf{w}$ and the twist of the slider $\mathbf{t}$. The applied wrench $\mathbf{w}$ causes the object to move in a perpendicular direction to the limit surface $\mathbf{H(w)}$. Consequently, the object twist in body frame is given by $\mathbf{t} = \nabla \mathbf{H(w)}$, where the applied wrench $\mathbf{w} = [f_{\overrightarrow{n}},f_{\overrightarrow{t}},\tau]$ could be written as $\mathbf{w} = \mathbf{J}^T(\overrightarrow{n} f_{\overrightarrow{n}} +\overrightarrow{t}f_{\overrightarrow{t}})$. For the contact configuration shown in Figure~\ref{fig:pushing_analytical}, the normal and tangential unit vectors are given by $\overrightarrow{n}=[1\quad 0]^T$ and $\overrightarrow{t}=[0\quad 1]^T$. 

The equations of motion of the pusher-slider system are
\begin{equation}
    \mathbf{\dot{x}}= \mathbf{f(\mathbf{x},\mathbf{u})}= \begin{bmatrix}
\mathbf{Rt}\quad
\dot{p}_y
\end{bmatrix}^\top
\label{eqn:pushing_dynamics}
\end{equation}
where $\mathbf{R}$ is the rotation matrix. Since the wrench applied on the system depends of the point of contact of pusher and slider, the state of the system is given by $\mathbf{x}= [x\quad y\quad \theta \quad p_y]^T$ and the input is given by $\mathbf{u} = [f_{\overrightarrow{n}}\quad f_{\overrightarrow{t}}\quad \dot{p}_y]^T$. The elements of the input vector must follow the laws of coulomb friction which can be expressed as complementarity conditions as follows:
\begin{align}
 0 \leq   \dot{p}_{y+} \perp (\mu_p f_{\overrightarrow{n}}(t)-f_{\overrightarrow{t}}(t)) \geq 0 \nonumber \notag\\
 0 \leq   \dot{p}_{y-} \perp (\mu_p f_{\overrightarrow{n}}(t)+f_{\overrightarrow{t}}(t)) \geq 0  
 \label{eqn:compl_pushing}
\end{align}

where $\dot{p}_y=\dot{p}_{y+}-\dot{p}_{y-}$ and the $\mu_p$ is the coefficient of friction between pusher and slider. The complementarity conditions in Eq.~\eqref{eqn:compl_pushing} mean that both $\dot{p}_{y+}$ and $\dot{p}_{y-}$ are non-negative and only one of them is non-zero at any time instant and $\dot{p}_y$ is non-zero only at the boundary of friction-cone. 

Two pushing trajectories with different goal configurations from the same initial state are shown in Figure~\ref{fig:example_pushing}. In both these examples, the initial pose of the slider is $\mathbf{x}_{\texttt{init}}=(0,0,0)$ and the desired goal pose of the slider is $\mathbf{x_g}=(0,0.5,\pi)$ and $(0,0,\pi)$. The initial point of contact between the pusher and the slider is $p_y=0$. For all these examples, the maximum normal force is set to $0.5 $ N and the coefficient of friction is $\mu_p=0.3$. The corresponding control trajectory shows the sequence of forces $f_n$ and $f_t$ used by the slider to obtain the desired trajectory. The plot of $\dot{p}_y$ shows the sequence of sticking and slipping contact as found by \pyrobocop\ and thus this also decides the contact point between the pusher and the slider. Note that the pusher maintains sticking contact with slider whenever $\dot{p}_y=0$, and slipping contact otherwise. %

\subsection{Assembly of Belt Drive Unit}
\begin{figure*}[!h]
     \centering
     \begin{subfigure}[b]{0.32\textwidth}
         \centering
         \includegraphics[width=\textwidth]{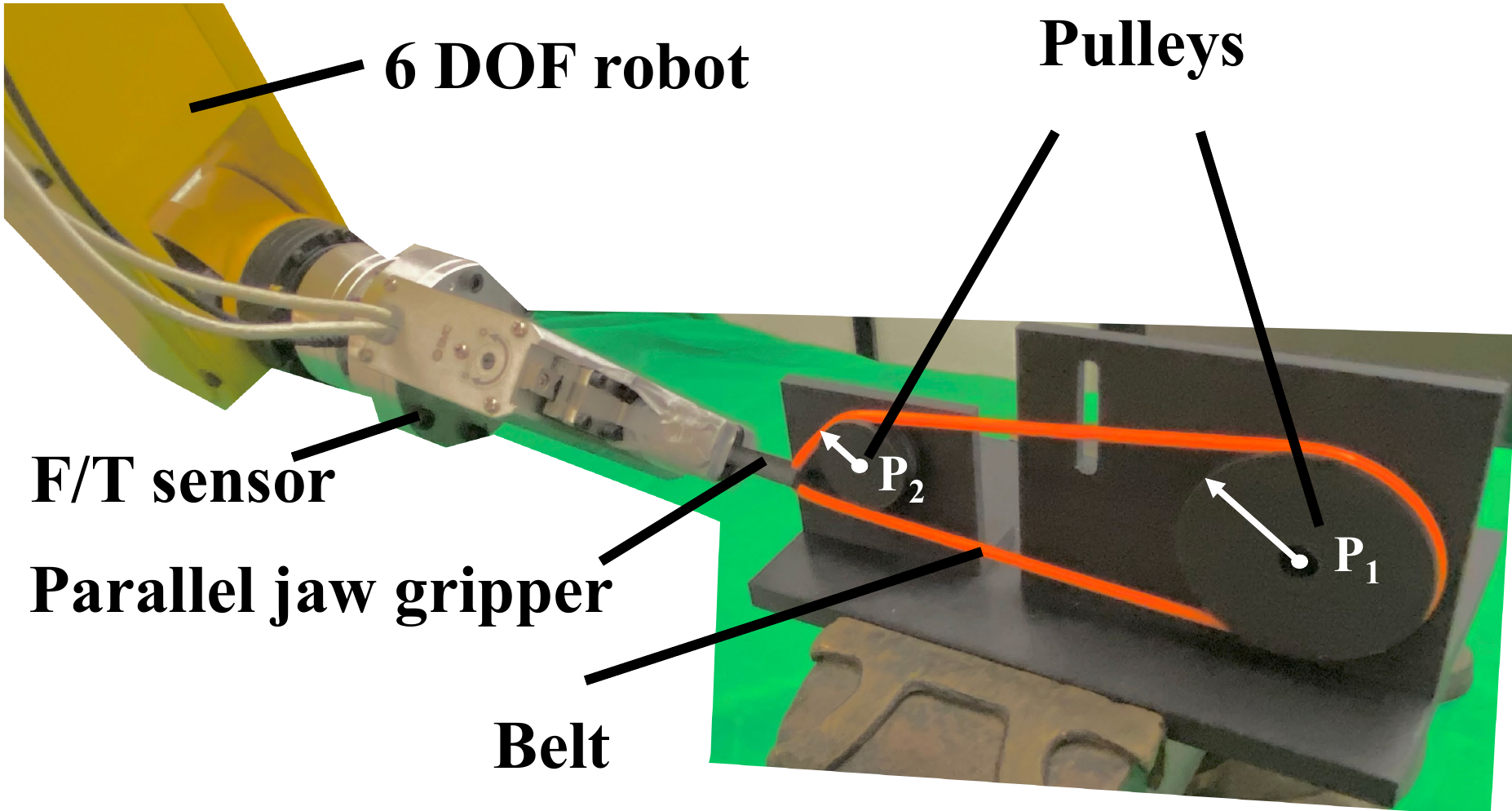}
    \caption{Real Setup of the Belt Drive Unit system.}
    \label{fig:BDU_setup}
     \end{subfigure}     
     \hfill
     \begin{subfigure}[b]{0.32\textwidth}
         \centering
         \includegraphics[width=\textwidth]{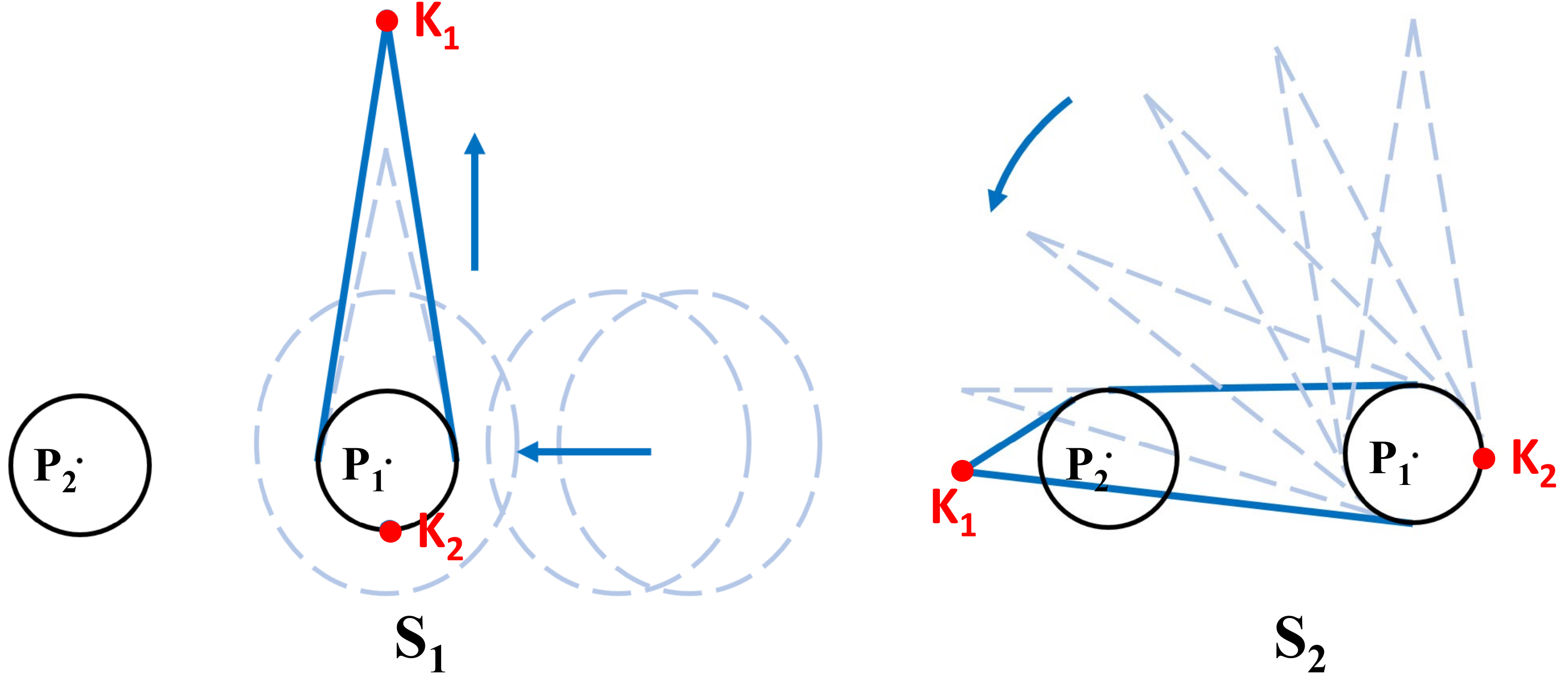}    \caption{Visualization of two subtasks decomposition, $S_1$ and $S_2$. $P_1$ and $P_2$ are the pulleys. %
         }
    \label{fig:BDU_subtasks}
     \end{subfigure}
          \hfill
     \begin{subfigure}[b]{0.32\textwidth}
         \centering
         \includegraphics[width=\textwidth]{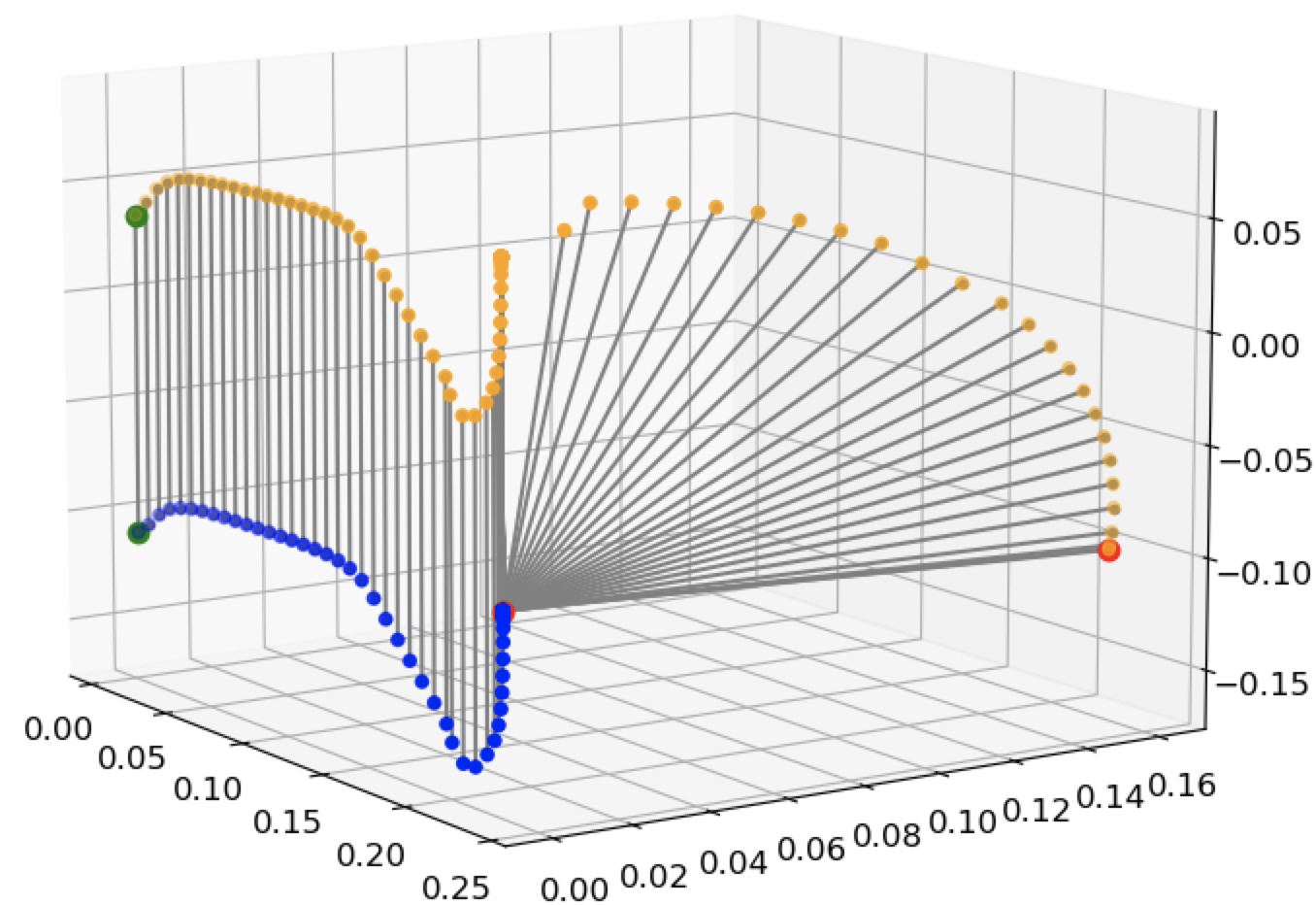}
    \caption{Optimal trajectory to assemble the belt. %
    }
    \label{fig:BDU_traj3D}
     \end{subfigure}
     \caption{BDU system is shown in (a). In (b) the blue lines represent the belt gripped  by a robot at keypoint $K_1$, and $K_2$ is the lower keypoint. In figure (c) the orange points represent the trajectory of the upper keypoint, $K_1$, and the blue points represent the lower keypoint, $K_2$, which together represent the model of the belt. The green and red points are the starting and the final points, respectively. The belt approaches the first pulley (not shown), then there is a downwards movement to hook the pulley with the lower keypoint during $S_1$. $K_1$ is now hooked onto the pulley and will not move. Then, the higher keypoint, $K_2$, moves toward the second pulley (not shown) stretching the belt and wraps around the pulley to conclude $S_2$.}
	\label{fig:BDU}
\end{figure*}

\begin{figure*}[ht]
     \centering
     \begin{subfigure}[b]{0.25\textwidth}
         \centering
         \includegraphics[width=\textwidth]{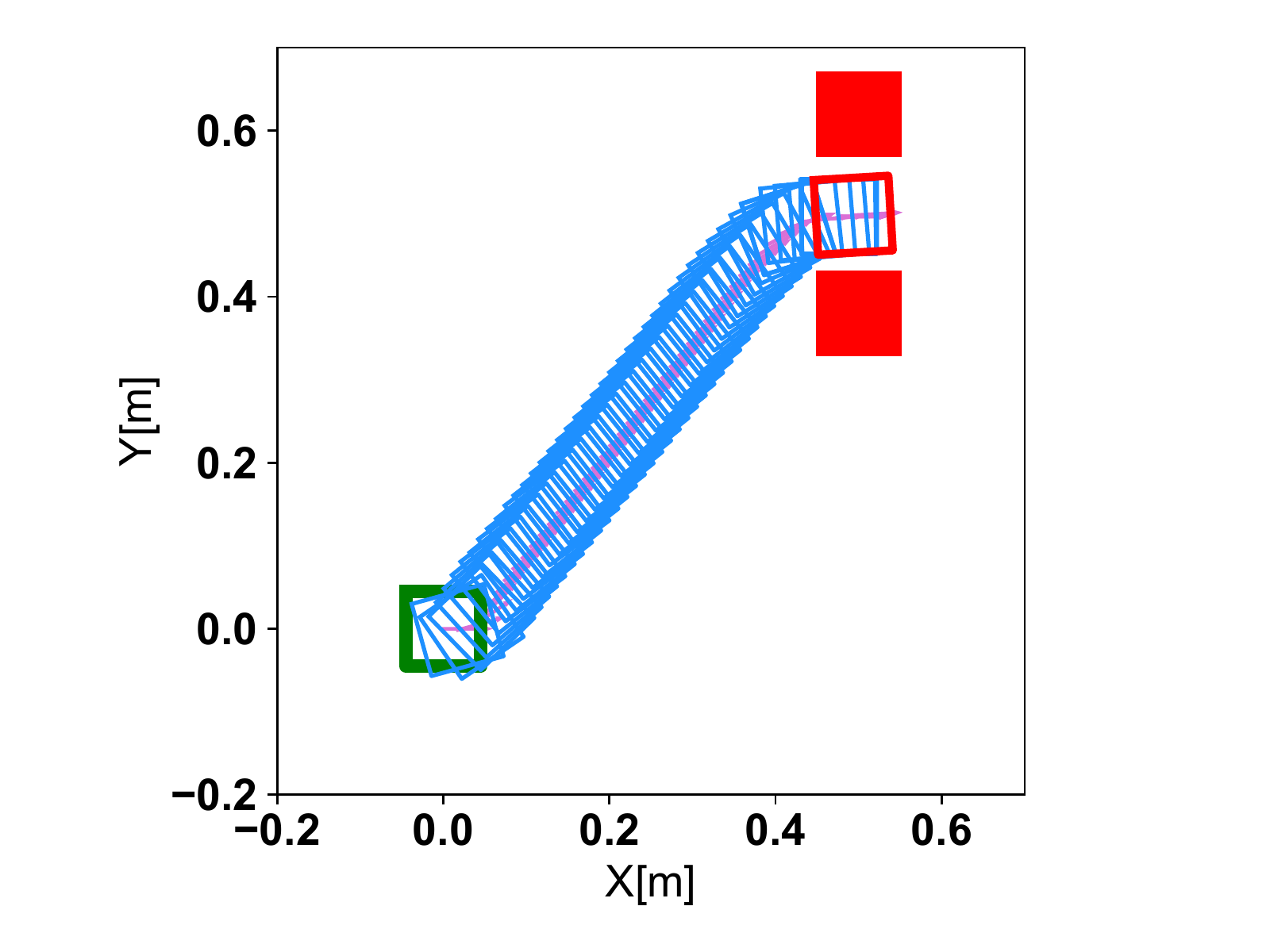}
		\vskip -0.05in
		\subcaption{Pushing sequence with~$\mathbf{x}_{\texttt{init}}$ $(0,0,0)$ and $\mathbf{x_g} = (0.5,0.5,0)$}\label{fig:pushing_obs_solution_ex1}
     \end{subfigure}
     \hfill
     \begin{subfigure}[b]{0.24\textwidth}
         \centering
         \includegraphics[width=\textwidth]{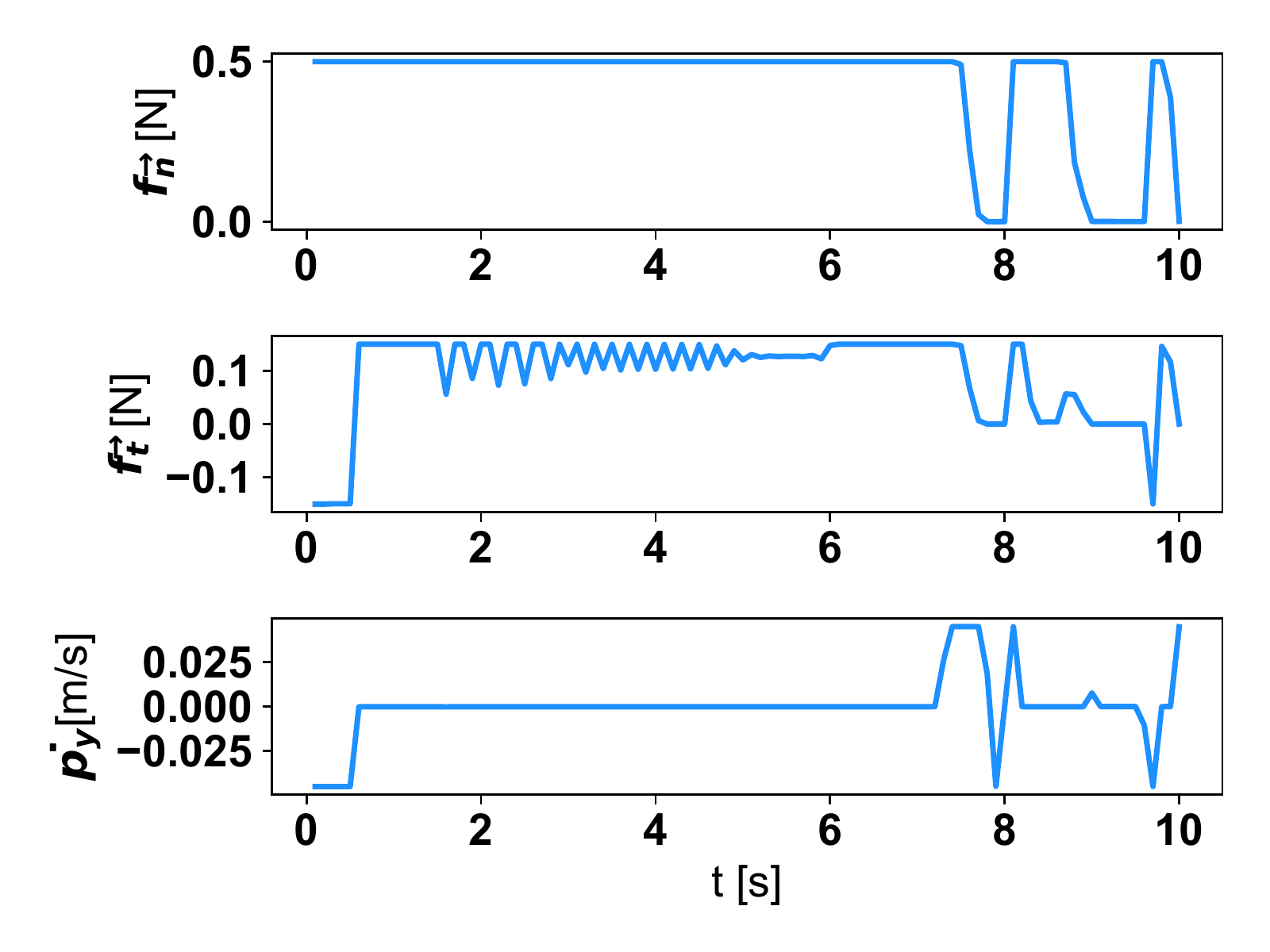}
		\vskip -0.05in
		\subcaption{Optimal Controls obtained for  Example~\ref{fig:pushing_obs_solution_ex1}}\label{fig:pushing_obs_solution_ex1_input}
     \end{subfigure}
     \hfill
          \centering
     \begin{subfigure}[b]{0.26\textwidth}
         \centering
         \includegraphics[width=\textwidth]{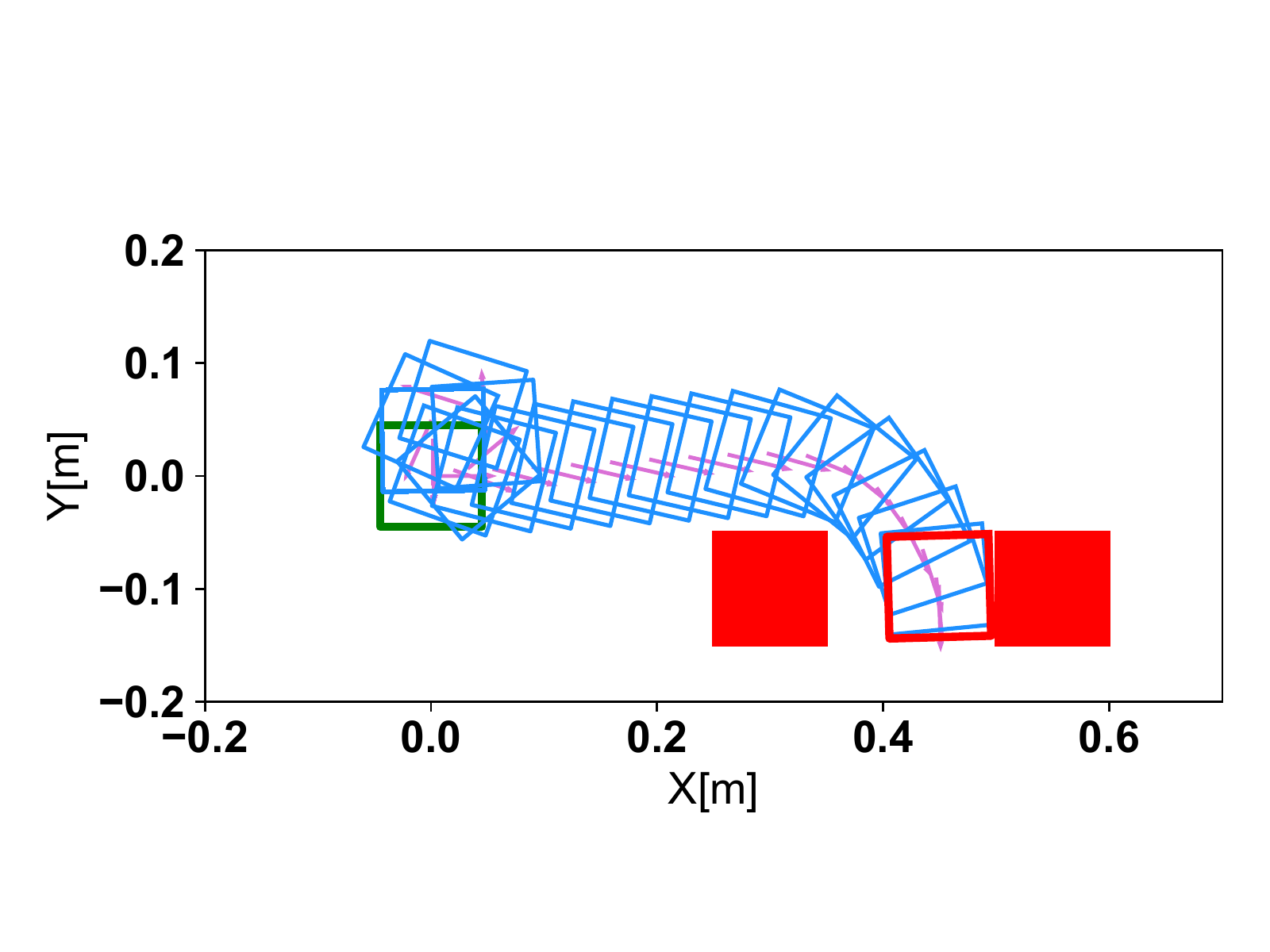}
		\vskip -0.05in
		\subcaption{Pushing sequence~$\mathbf{x}_{\texttt{init}}$ $(0,0,0)$ and~$\mathbf{x_g} =(0.45,-0.1,\frac{3\pi}{2})$}\label{fig:pushing_obs_solution_2}
     \end{subfigure}
     \hfill
     \begin{subfigure}[b]{0.23\textwidth}
         \centering
         \includegraphics[width=\textwidth]{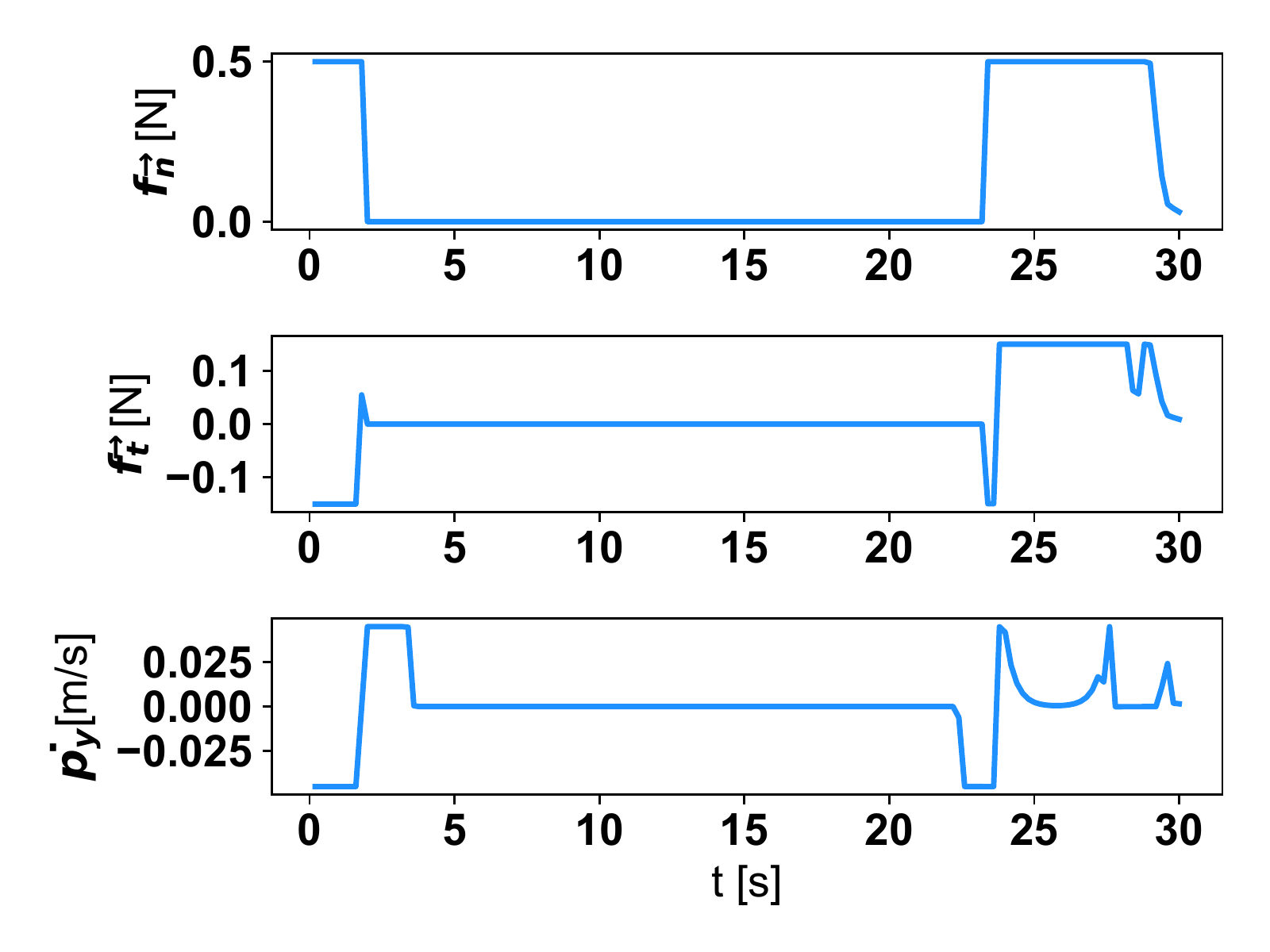}
		\vskip -0.05in
		\subcaption{Optimal Controls obtained for Example~\ref{fig:pushing_obs_solution_2}.}\label{fig:pushing_obs_solution_2_input}
     \end{subfigure}
     \caption{Planar pushing in the presence of obstacles. Our proposed formulation in \pyrobocop\ allows us to solve the collision avoidance. The trajectory of $\dot{p}_y$ shows the slipping contact sequence between the pusher and the slider. Pusher maintains a sticking contact when $\dot{p}_y=0$.}
	\label{fig:example_pushing_with_obstacles}
\end{figure*}

An example of a complex manipulation problem that involves contacts, elastic objects and collision avoidance is provided by the Belt Drive Unit system. This assembly challenge was presented as one of the most challenging competition in the World Robot Summit 2018\footnote{\url{https://worldrobotsummit.org/en/about/}}~\cite{drigalski2019}.  
The real world system is represented in Figure~\ref{fig:BDU_setup} where the objective of the manipulation problem is to wrap the belt, held by a robotic manipulator around the two pulleys. The elastic belt is modeled through a 3D keypoint representation. The hybrid behavior of the model generated by the contacts between the belt and the pulleys and the elastic properties of the belt is captured by the complementarity constraints. 

The full manipulation task has been divided into two subtasks as shown in Figure~\ref{fig:BDU_subtasks}. The goals of the first and second subtask are to wrap the belt around the first and second pulley, respectively. 
We formulate two trajectory optimization problems one for each of the two subtasks as a MPCC of the form in~\eqref{dynopt}.

Details on the modeling assumptions, the division into the two subtasks, the exact formulation including the explanation of the dynamics and complementarity constraints can be found in our previous paper~\cite{jin2021trajectory}. In Figure~\ref{fig:BDU_traj3D} we report successful trajectories computed by \pyrobocop\  to assemble the belt drive unit combining the optimal trajectories obtained in the two subtasks. The optimal trajectory was implemented on the real system with a tracking controller, see~\cite{jin2021trajectory} for further details.

\subsection{Planar Pushing With Obstacles}

In this section, we show the solution to some planar pushing scenarios in the presence of obstacles and show that our proposed method can handle complementarity constraints as well as obstacle avoidance constraints simultaneously. We demonstrate our approach on two different pushing scenarios with same initial condition for the slider but different location of the obstacles and different goal states for the slider. In particular, the initial state of the slider in both these examples was set to $\mathbf{x}_{\texttt{init}}=(0,0,0)$ and the goal state for the two conditions was specified as $\mathbf{x}_g = (0.5,0.5,0)$ and $(-0.1,-0.1,3\pi/2)$. We add the obstacles next to the goal state so that \pyrobocop\ has to find completely different solution compared to the case when there are no obstacles. The initial point of contact between the pusher and the slider is $p_y=0$.
The optimal pushing sequence to reach the goal states for the slider are shown in  Figures~\ref{fig:pushing_obs_solution_ex1} and~\ref{fig:pushing_obs_solution_2}. To provide more insight about the solution, we also provide the plot of the input sequences in Figures~\ref{fig:pushing_obs_solution_ex1_input} and~\ref{fig:pushing_obs_solution_2_input}. The slipping contact sequence between the slider and the pusher is seen in the plot of $\dot{p}_y$. Sticking contact occurs when $\dot{p}_y=0$. We show that the proposed solver can optimize for the desired sequence of contact modes in order to reach the target state. For the example in Figure~\ref{fig:pushing_obs_solution_ex1}, the objective is a function of target state and control inputs. For the example in Figure~\ref{fig:pushing_obs_solution_2}, the Mayer objective function is used.

\subsection{Optimization with Mode Enumeration}

\begin{figure}
        \centering
        \begin{subfigure}[b]{0.49\columnwidth}
            \centering
            \includegraphics[width=\textwidth]{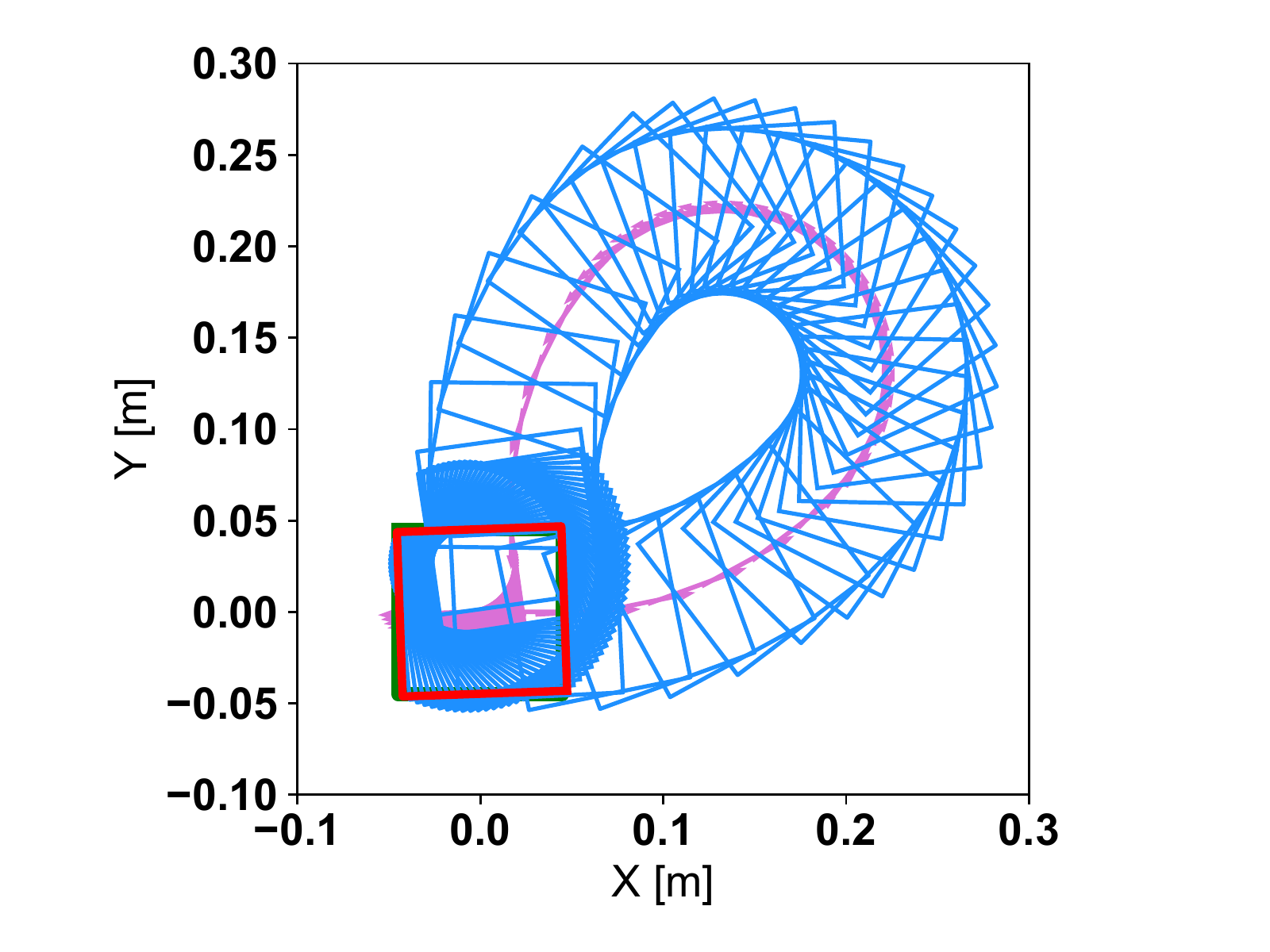}
            \caption{Optimal pushing sequence.}
            \label{fig:pushing_mode_seq}
        \end{subfigure}
        \hfill
        \begin{subfigure}[b]{0.49\columnwidth}  
            \centering 
            \includegraphics[width=\textwidth]{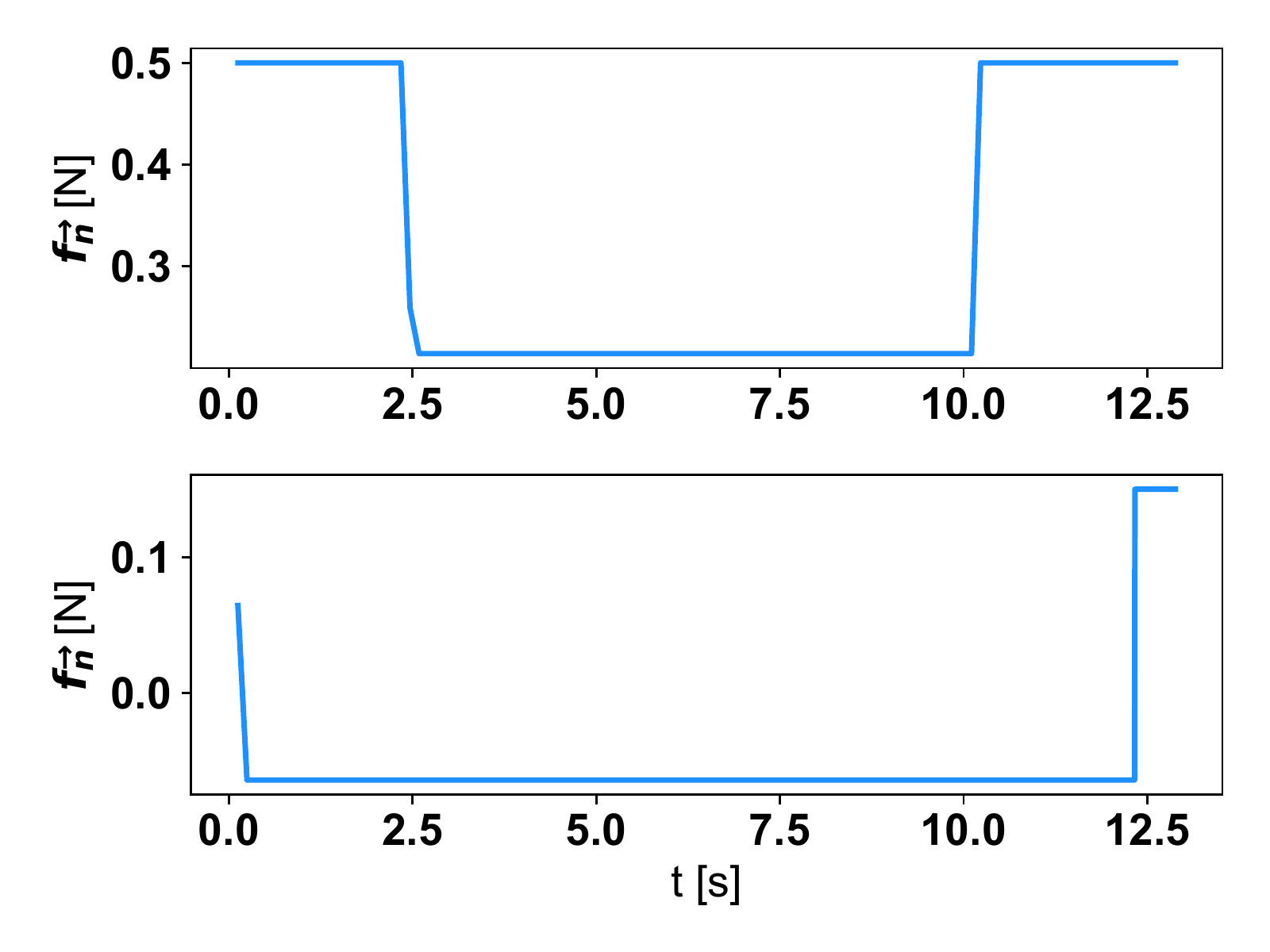}
            \caption{Optimal control inputs.}
            \label{fig:pushing_mode_seq_inputs}
        \end{subfigure}
        \caption{Optimal pushing sequence and control inputs obtained by optimizing mode sequence with \pyrobocop.}
        \label{fig:mode_seq_pushing}
\end{figure}

We show our optimization approach over fixed mode sequences using the quasi-static pushing model presented in Section~\ref{subsec:planar_pushing} while considering sticking contact at the $4$ faces of the slider (see Figure~\ref{fig:pushing_analytical}). In particular, we use the dynamics model and the problem described in~\cite{doshi2020hybrid} to show solutions obtained by \pyrobocop\ in the case the mode sequence is pre-specified and the resulting problem is feasible. %

The contact model in this case can be obtained from the model described in Section~\ref{subsec:planar_pushing}, Eq~\ref{eqn:pushing_dynamics} with $\dot{p}_y=0$. The modes appear based on which face the pusher contacts with the slider, and thus we have four different modes that could be used during any interaction. For a given mode, state-space of the pusher-slider system is then $3$ dimensional while the input is only $2$ dimensional. The optimization process ensures continuity of dynamics and selection of final time for each mode in a trajectory. The initial state of the slider is $\mathbf{x}_{\texttt{init}}= (0,0,0)$ and the goal state of the slider is $\mathbf{x_g}=(0,0,\pi)$. The two modes we use for this example are pushing from the left face followed by pushing from the top face of the slider. The trajectory obtained by \pyrobocop\ is shown in Figure~\ref{fig:pushing_mode_seq}. The inputs used in different modes is shown in Figure~\ref{fig:pushing_mode_seq_inputs}. The objective function used is the Mayer objective function, i.e., the minimum-time problem. The time spent in mode $1$ is $12.36$[s] and in mode $2$ is $0.56$[s].

\subsection{System Identification For Complementarity Systems}
The system identification problem for systems with complementarity constraints is a particular case of the optimization problem \eqref{nlp} and therefore can be solved in \pyrobocop. The objective is to identify the physical parameters of a system given a set of collected data. As a case of study, we consider the cart-pole with softwalls depicted in Figure~\ref{fig:pushing_analytical}. The interactions of the pole with the soft walls is modeled as complementarity constrains. The dynamical equations and more details on the system can be found in~\cite{aydinoglu2020stabilization}. 

\begin{figure}
    \centering
    \includegraphics[scale=0.50]{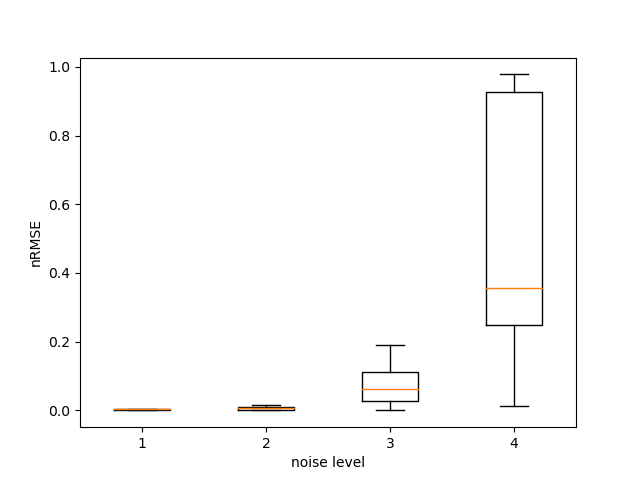}
    \caption{Distributions of the estimation errors for 4 different cart-pole with softwalls with increasing noise level.}
    \label{fig:sysid_err}
\end{figure}

We formalize the parameter estimation problem as MPCC \eqref{nlp} where the cost function is the normalized Root Mean Square Error (nRMSE) between the observed trajectory and the trajectory obtained from the estimation procedure. The parameters we aim to identify are the mass of the pole, $m_p$, and the spring constants of the two walls $k_1$ and $k_2$. In \pyrobocop\, these parameters are implemented as time independent parameters $p$. We validated the method with a Monte Carlo simulation on 4 different sets of parameters $p^i = [m_p^i,k_1^i,k_2^i]$ with $i=\{1,\ldots,4\}$ sampled independently from a uniform distribution, each of which was tested with different levels of independent Gaussian noise added to the trajectories collected. The trajectories are generated with an input sequence that is computed as a sum of sinusoids. Each MC simulation has 50 random realization of the noise. The results are shown in Figure~\ref{fig:sysid_err} where on the x-axis we have the cart-pole system defined with one of the parameter set $p^i$ and the standard deviation of the noise for each system is in order $[0.0001,0.001,0.01,0.05]$.
We can observe how \pyrobocop\ is able to identify both the parameters in the dynamics equations as well as in the complementarity constraints with the lower levels of noise. As seen in the figure, we observe that with higher amounts of noise, the estimation method starts diverging.

\section{Concluding Remarks}\label{sec:conclusions}
This paper presented \pyrobocop\,, a Python-based optimization package for model-based control of robotic systems. We showed that \pyrobocop\ can be used to solve trajectory optimization problems of a number of dynamical systems in different configurations in presence of contact and collision avoidance constraints. A description of the functions that a potential user needs to implement in order to solve their control or optimization problem has been provided. More details are available in the software package~\cite{pyrobocop_software}.%

\textbf{Strengths of \pyrobocop}
\pyrobocop\ can handle systems with contact as well as collision constraints with a novel complementarity formulation. \pyrobocop\ also allows automatic differentiation by using \adolc. To the best of our knowledge, \pyrobocop\ is the only Python-based, open-source software that allows handling of contact \& collision constraints and automatic differentiation for control and optimization. Unlike most of the competing optimization solvers which are available in Python, \pyrobocop\ allows users to provide dynamics information in Python through a simple script, refer to the software description in~\cite{pyrobocop_software}. %

\textbf{Limitations of \pyrobocop\ }  Since \pyrobocop\ uses \ipopt\ as the solver for the resulting MPCC problems, it borrows limitations of \ipopt. In particular, one of the main limitations is that only local solutions can be found. Furthermore, good initialization to find even the local solutions might be required. Finally, infeasibility of the underlying optimization problem provided by the user cannot be detected. Another possible limitation is given by interfacing \pyrobocop\ with \adolc. While, as described above, this is one of the strengths of \pyrobocop\, it also carries some limitations as we have to rely on an external code to do automatic differentiation, while other software like \casadi\ have built-in source code transformation into C and can handle the differentiation internally with faster performance. %

\bibliographystyle{IEEEtran}
\bibliography{references}

\end{document}